\documentclass[lettersize,journal]{IEEEtran}
\usepackage{amsmath,amsfonts}
\usepackage{algorithmic}
\usepackage{algorithm}
\usepackage{array}
\usepackage[caption=false,font=normalsize,labelfont=sf,textfont=sf]{subfig}
\usepackage[pagebackref=true,breaklinks=true,colorlinks,bookmarks=false]{hyperref}
\usepackage{textcomp}
\usepackage{stfloats}
\usepackage{url}
\usepackage{verbatim}
\usepackage{graphicx}
\usepackage{cite}

\usepackage{xcolor} 
\usepackage{booktabs}
\usepackage{multirow}

\begin{document}

\title{CLIP-VIS: Adapting CLIP for Open-Vocabulary Video Instance Segmentation}

\author{Wenqi Zhu, Jiale Cao, Jin Xie, Shuangming Yang, and Yanwei Pang
\thanks{This work was supported by the National Key Research and Development Program of China (No. 2022ZD0160400), and the Natural Science Foundation of China (No. 62271346, 62206031). \textit{(Corresponding author: Jiale Cao)}}
\thanks{W. Zhu, J. Cao, S. Yang, and Y. Pang are with the School of Electrical and Information Engineering, Tianjin University, Tianjin 300072, China, also with Shanghai Artificial Intelligence Laboratory, Shanghai 200232, China  (E-mail: \{zhuwenqi, connor, yangshuangming, pyw\}@tju.edu.cn).}
\thanks{J. Xie is with the School of Big Data and Software Engineering, Chongqing University, Chongqing 401331, China (E-mail: xiejin@cqu.edu.cn).}}

\markboth{Journal of \LaTeX\ Class Files,~Vol.~14, No.~8, 20~21}
{ZHU \MakeLowercase{\textit{et al.}}: CLIP-VIS: Adapting CLIP for Open-Vocabulary Video Instance Segmentation.}

\IEEEpubid{\begin{minipage}{\textwidth}\centering
Copyright \copyright~20xx IEEE. Personal use of this material is permitted. \\However, permission to use this material for any other purposes must be obtained from the IEEE by sending an email to pubs-permissions@ieee.org.
\end{minipage}}
\maketitle

\begin{abstract}
Open-vocabulary video instance segmentation strives to segment and track instances belonging to an open set of categories in a videos. The vision-language model Contrastive Language-Image Pre-training (CLIP) has shown robust zero-shot classification ability in image-level open-vocabulary tasks. In this paper, we propose a simple encoder-decoder network, called CLIP-VIS, to adapt CLIP for open-vocabulary video instance segmentation. Our CLIP-VIS adopts frozen CLIP and introduces three modules, including class-agnostic mask generation, temporal top$K$-enhanced matching, and weighted open-vocabulary classification. Given a set of initial queries, class-agnostic mask generation introduces a pixel decoder and a transformer decoder on CLIP pre-trained image encoder to predict query masks and corresponding object scores and mask IoU scores. Then, temporal top$K$-enhanced matching performs query matching across frames using the $K$ mostly matched frames. Finally, weighted open-vocabulary classification first employs mask pooling to generate query visual features from CLIP pre-trained image encoder, and second performs weighted classification using object scores and mask IoU scores. Our CLIP-VIS does not require the annotations of instance categories and identities. The experiments are performed on various video instance segmentation datasets, which demonstrate the effectiveness of our proposed method, especially for novel categories. When using ConvNeXt-B as backbone, our CLIP-VIS achieves the AP and AP$_n$ scores of 32.2\% and 40.2\% on the validation set of LV-VIS dataset, which outperforms OV2Seg by 11.1\% and 23.9\% respectively. We will release the source code and models at \url{https://github.com/zwq456/CLIP-VIS.git}.
\end{abstract}


\section{Introduction}
\IEEEPARstart{V}{ideo} instance segmentation (VIS)\cite{yang2019video} aims to simultaneously segment and track the instances in a video. The related video instance segmentation methods can be divided into online \cite{yang2019video,huang2022minvis,wu2022defense} and offline \cite{wang2021end,wu2022seqformer,heo2022vita} approaches. The online approaches first segment the instances existing in each frame, and then associate the instances across frames. Different from online approaches, offline approaches explore spatial-temporal information within a video to directly segment and associate the instances across frames.  In past few years, video instance segmentation has achieved promising progress. However, these traditional video instance segmentation approaches usually assume that the training and test sets share  the same pre-defined closed set of instance categories, which restricts their application in real-word open environments.

\begin{figure}
\centering
\includegraphics[width=1.0\linewidth]{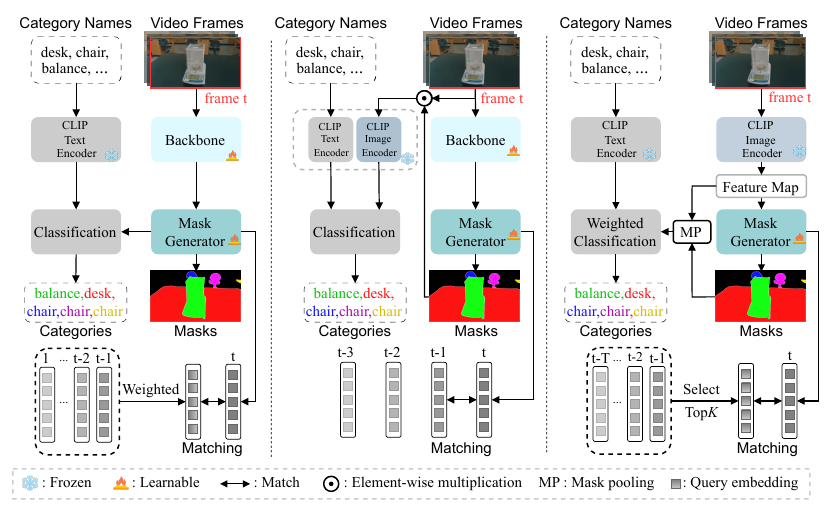}
\caption{\textbf{Comparisons with existing open-vocabulary video instance segmentation pipelines.} The inputs are category names and video frames, and the outputs are  masks, categories and matching results across frames of different instances.\textit{Left}: OV2Seg employs visual features of trainable backbone and  CLIP text encoder for open-vocabulary classification. Instance query matching is performed using all previous (long-term) query embeddings. \textit{Middle}: OpenVIS first utilizes a trainable network to extract proposal masks, and then employs frozen CLIP for classification. Instance query matching is conducted using adjacent query embeddings. \textit{Right}: Our CLIP-VIS performs mask generation and classification using a single frozen CLIP. Query matching is performed based on query embeddings of $K$ mostly matched frames.} 
\label{meta_arch}
\end{figure}

Recently, the researchers \cite{wang2023towards,guo2023openvis} started to explore open-vocabulary video instance segmentation tasks with the goal of segmenting and tracking instances belonging to arbitrary object categories.  As in Fig. \ref{meta_arch}, OV2Seg \cite{wang2023towards} first employs a mask proposal network to extract class-agnostic masks and second performs open-vocabulary classification using a CLIP \cite{radford2021learning} text encoder. Afterwards, OV2Seg conducts instance query matching using all previous frames with a long-term matching strategy. In contrast, 
OpenVIS \cite{guo2023openvis} initially extracts masks across frames, and then employs a CLIP model to classify each cropped instance from the input image. Then, OpenVIS performs query matching using one adjacent frame. 
\IEEEpubidadjcol
Though these two approaches present the initial attempt at open-vocabulary video instance segmentation task, we argue that they have some limitations: (i) OV2Seg directly aligns visual features of trainable network and text embeddings of frozen CLIP text encoder, which is prone to overfit the categories of the training set. In addition, OV2Seg performs the  long-term matching  using all previous frames that easily incorporate noisy frames. (ii) OpenVIS is not an end-to-end approach, which requires two independent networks, including a trainable network and a frozen CLIP, for instance segmentation and classification. Additionally, its adjacent matching strategy easily propagates matching errors from the current frame to subsequent frames.

To address their limitations, we propose a simple network to directly adapt CLIP model  \cite{radford2021learning} for open-vocabulary video instance segmentation, called CLIP-VIS.By pre-training on large-scale image-text dataset, CLIP has shown strong open-vocabulary classification capabilities across numerous vision tasks. Therefore, we adopt the encoder-decoder structure as in Mask2Former  \cite{cheng2022masked} and replace the original visual backbone by frozen CLIP image encoder to fully utilize its strong zero-shot classification ability. To generate pixel-level features, we remove the pooling operation in original CLIP image encoder. Based on the modified encoder-decoder, we introduce three modules, including  class-agnostic mask generation, temporal top$K$-enhanced matching, and weighted open-vocabulary classification. Given a set of initial queries, the class-agnostic mask generation predicts query masks for each frame along with their corresponding object scores and mask IoU scores. To achieve this goal, class-agnostic mask generation attaches a mask generator, including a pixel decoder and a transformer decoder, on CLIP pre-trained image encoder. Class-agnostic mask generation does not require class supervision during training, which is particularly beneficial in open real-world settings, where it is impractical and expensive to annotate numerous object categories. Subsequently, the temporal top$K$-enhanced matching selects $K$ mostly matched frames from previous frames and performs bipartite matching based on query embeddings of current frames and query embeddings from these selected frames. This strategy effectively mitigates matching error propagation and reduces interference from noisy frames compared to using solely adjacent or long-term frames. Finally, inspired by ODISE \cite{xu2023open} and FC-CLIP\cite{yu2024convolutions}, weighted open-vocabulary classification employs mask pooling to generate query visual features from CLIP pre-trained image encoder, and calculates classification scores according to the similarity of query visual features and CLIP text embeddings that generated by CLIP pre-trained text encoder. To better correlate mask predictions and classifications, our weighted open-vocabulary classification refines classifications with object scores and mask IoU scores. We  evaluate the performance  on various datasets, including LV-VIS \cite{wang2023towards}, YTVIS2019/2021 \cite{yang2019video}, OVIS \cite{qi2022occluded}, and BURST \cite{athar2023burst}.  The contributions and merits  can be summarized as follows.

\begin{itemize}
\item{We propose  to adapt frozen CLIP for  open-vocabulary video instance segmentation. Our method retains strong zero-shot classification ability  to various categories.}
\item{We design a temporal top$K$-enhanced  matching strategy to adaptively select $K$ mostly  frames for query matching.}
\item{We further introduce a weighted open-vocabulary classification module, which refines mask classification by correlating mask prediction and classification.}
\item{Our CLIP-VIS achieves superior performance on  multiple datasets. On LV-VIS validation set, CLIP-VIS outperforms OV2Seg by 5.3\% and 11.1\% $AP$ using  R50 and ConvNext-B, respectively. Furthermore,  on novel categories, CLIP-VIS achieves a significant improvement, outperforming OV2Seg by 12.3\% and 23.9\% $AP$, which demonstrates the robust zero-shot classification ability for unseen categories. With R50, CLIP-VIS outperforms OpenVIS by 1.6\% $AP$ and 1.4\% $AP_u$ on BURST validation set, further demonstrating its effectiveness.}
\end{itemize}

\section{Related Work}

In this section, we first introduce classical video instance segmentation. Afterwards, we give a brief review on vision-language models. Finally, we introduce open-vocabulary segmentation methods based on vision-language models.

\subsection{Video Instance Segmentation} 

Video instance segmentation (VIS) is derived from image instance segmentation \cite{10203974,10048550,9367228}, video object segmentation\cite{wang2023look,xi2023online,chen2023boosting}, and multiple object tracking\cite{qin2023motiontrack,8766896,zhang2023motrv2}, presenting greater challenges. Existing methods can be primarily categorized into two distinct classes, including online \cite{fu2021compfeat,han2022visolo,he2022inspro} and offline \cite{yang2022temporally,wu2022efficient,fang2024learning} approaches.

Online approaches tackle VIS by first performing image instance segmentation on individual frames and subsequently associating instances across frames. MaskTrack R-CNN \cite{yang2019video} is one of the earliest VIS approaches, which adds an instance tracking branch to  Mask R-CNN\cite{he2017mask}. In tracking branch, MaskTrack R-CNN stores the instance embeddings of previous frame in memory, and performs instance matching between instance embeddings of current frame and that in memory. Afterwards, some singe-stage approaches were proposed built on single-stage image framework \cite{bolya2019yolact,tian2020conditional,yang2021crossover}. For example, SipMask \cite{cao2020sipmask} extends single-stage instance segmentation approach YOLACT \cite{bolya2019yolact} by adding a fully-convolutional tracking branch. CrossVIS \cite{yang2021crossover} extends single-stage approach CondInst \cite{tian2020conditional} and improves performance by performing cross-frame tracking feature learning, which exchanges the features of same instance across frames to perform mask generation during training. Recently, with the success of Transformer \cite{vaswani2017attention}, some transformer-based approaches were proposed. For example, MinVIS \cite{huang2022minvis} observes that the same instance embeddings generated by transformer decoder have high similarity across different frames. Therefore, MinVIS directly employs  Hungarian algorithm \cite{kuhn1955hungarian} for instance association across frames. IDOL\cite{wu2022defense} introduces a contrastive learning strategy to make the embeddings of different instances within a frame more dissimilar and the embeddings of same instance across different frames more similar. CTVIS\cite{ying2023ctvis} points out that the contrastive learning strategy in IDOL only compares instance embeddings between two frames, resulting in less stable instance representations. Therefore, CTVIS introduces multi-frame contrastive learning to make the instance embeddings more representative. TCOVIS \cite{li2023tcovis} introduces global instance assignment between predictions and video ground-truths during training. 

In contrast to online methods, offline approaches directly segment and associate the instances in a complete video sequence. VisTR\cite{wang2021end} is one of typical offline approaches built on a transformer structure. VisTR views the VIS task as a set prediction problem, and employs transformer to predict the masks of instances within a video. SeqFormer\cite{wu2022seqformer} first divides the instance queries into frame-level instance queries to perform feature extraction at each frame, and then combines these frame-level queries for mask prediction and classification. 
Similarly, VITA\cite{heo2022vita} first generate frame-level query features. Afterwards, VITA generate video-level query features for mask prediction by performing temporal interactions between frame-level queries. GenVIS\cite{heo2023generalized}  stores object queries in previous clips to guide feature learning in current clip, and allows flexible transitions between offline, semi-offline, and online modes.

Video instance segmentation approaches mentioned above assume that the training and inference stages share a same pre-defined closed set of instance categories. As a result, these methods are inherently constrained in their capacity to accurately segment and track instances of novel categories, thus limiting their practical applicability in real-world scenarios.

\subsection{Vision-Language Models}

Vision-language models \cite{lu2019vilbert,tan2019lxmert,chen2020uniter} are designed to learn the correlations between image-text pairs. CLIP \cite{radford2021learning} is one of most representative vision-language models, comprising an image encoder for extracting visual features from images and a text encoder for generating textual embeddings from text descriptions. CLIP calculates the cosine similarity between visual and text embeddings for image classification. To enhance feature representation, CLIP is trained on a large-scale dataset of image-text pairs collected from website. CLIP exhibits robust zero-shot transfer ability over various classification tasks. However, it is time-consuming to acquire a high-quality, large-scale image-text paired dataset. FILIP \cite{yao2021filip} further enhances this learning process by introducing a cross-modal fine-grained interaction mechanism, leveraging token-wise maximum similarity between visual and textual tokens to guide the contrastive objective. GLIP\cite{li2022grounded} introduces phrase grounding to establish fine-grained connections between image regions and textual phrases. This model executes language-aware deep fusion across multiple encoder layers, thereby enhancing features from both image and text modalities. In this paper, we aim to exploit the robust zero-shot transfer ability of CLIP for open-vocabulary classification, and extend CLIP for mask prediction and instance association.

\subsection{Open-Vocabulary Segmentation}
Compared to  the classical segmentation task that focuses on fixed category set, open-vocabulary segmentation task aims to segment objects or instances of arbitrary categories in images or videos. Most open-vocabulary segmentation methods leverage vision-language models \cite{radford2021learning,yao2021filip,li2022grounded} to perform the classification for open-vocabulary segmentation. The early works\cite{ding2022decoupling,xu2022simple,liang2023open} mainly adopt a two-stage pipeline. The first stage typically involves generating class-agnostic mask proposals, while the second stage employs a pre-trained CLIP model to perform mask classification. For instance, MaskCLIP \cite{ding2023maskclip}  generates class-agnostic mask proposals, and then fine-tunes CLIP image encoder to better align with segmentation task. Similarly, MasQCLIP \cite{xu2023masqclip}  employs distillation learning to enhance the performance of mask generation in the first stage, and incorporates MasQ-Tuning fine-tuning strategy to adjust the CLIP image encoder for classification in second stage. However, the aforementioned two stage methods suffer from redundancy due to the utilization of separate networks for mask generation and classification, leading to duplicate image feature extraction. To address this issue, several single-stage networks\cite{li2022languagedriven,xu2023side,xie2023sed} were introduced. ZegCLIP \cite{zhou2023zegclip} adapts CLIP for open-vocabulary segmentation by matching text and local visual embeddings from CLIP. FC-CLIP \cite{yu2024convolutions} utilizes CLIP image encoder for both mask generation and classification. Additionally, it integrates the scores of in-vocabulary and out-vocabulary classifiers to enhance classification performance. Our proposed method is related to FC-CLIP, but our network does not require category labels during training. 

Recently, the researchers started to explore open-vocabulary video instance segmentation task {\cite{wang2023towards,guo2023openvis,cheng2024instance,fang2024unified}}. OV2Seg \cite{wang2023towards} first extracts class-agnostic masks and corresponding query embeddings based on Mask2Former \cite{cheng2022masked}, and then employs the CLIP text encoder to classify  these masks. To track instances across frames, OV2Seg proposes to perform long-term matching using query embeddings of all previous frames. Similar to OV2Seg, OpenVIS \cite{guo2023openvis} first extracts class-agnostic masks. Afterwards, OpenVIS uses these masks to crop the instances from input frames, and employs the frozen CLIP to classify these cropped instances. Finally, OpenVIS tracks the instances by performing adjacent matching with one previous frame. BriVIS \cite{cheng2024instance} 
views frame-level query features as a Brownian bridge, and proposes to align the bridge
center with the category text embeddings for better open-vocabulary classification. BriVIS also adopts adjacent matching as in OpenVIS. OVFormer\cite{fang2024unified} introduces a new unified embedding alignment module to address the domain gap between instance queries and text embeddings in OV2Seg. Additionally, OVFormer adopts video-based training. Although these methods have achieved initial success in open-vocabulary video instance segmentation, there are some limitations: (1) OV2Seg directly aligns the visual features of conventional trainable network and text embeddings of CLIP pre-trained text encoder, which is easy to overfit the  categories used in training. (2) OpenVIS adopts two independent networks for mask generation and classification, which is  redundant and not optimal. (3) BriVIS and OVFormer are offline or semi-online approach, which limits its application on dealing with long or streaming videos. (4) The long-term matching in OV2Seg easily incorporates noisy frames, while adjacent matching in OpenVIS and BriVIS suffers from matching error propagation. In this paper, we introduce a simple and online framework by adapting CLIP for open-vocabulary video instance segmentation. To improve matching, we introduce a  temporal top$K$-enhanced matching strategy.

\begin{figure*}[t]
	\centering
	\includegraphics[width=1.0\textwidth]{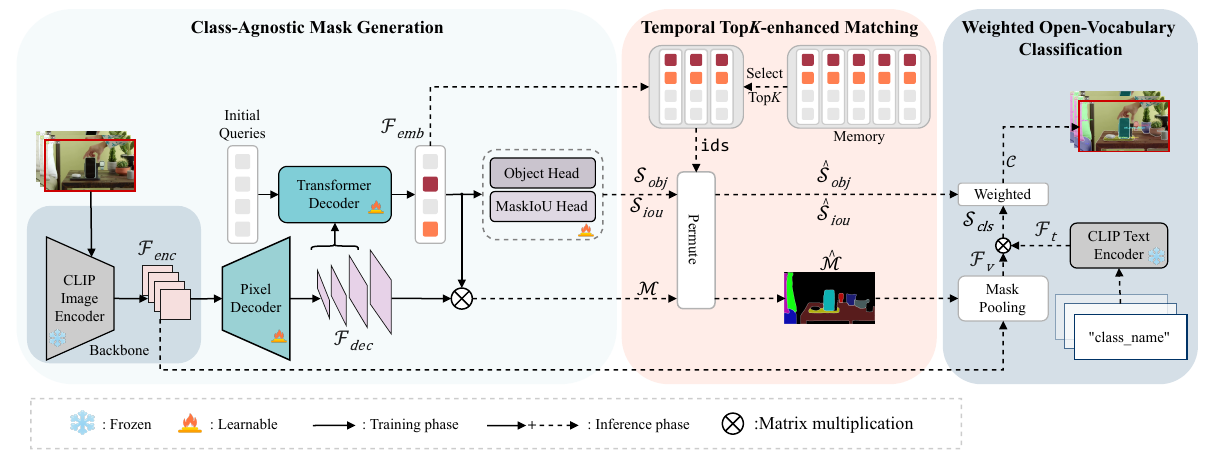} %
	\caption{\textbf{Overall architecture of our proposed CLIP-VIS.} It mainly comprises three  modules: \textit{a) Class-agnostic mask generation.} It employs CLIP image encoder as backbone, and  utilizes pixel decoder and transformer decoder in Mask2former to extract query masks. In addition, it generates mask IoU scores and object scores. \textit{b) Temporal top$K$-enhanced matching.} This module selects $K$ mostly matched frames and  associates query embeddings between current frame and these selected frames. \textit{c) Weighted open-vocabulary classification.} This module extracts query visual features from CLIP backbone using mask pooling, and calculates similarity scores between  query visual features and CLIP text embeddings for classification. In addition, we refine the classification by object scores and mask IoU scores. }
	\label{overview}
\end{figure*}

\section{Method}
In this section, we introduce our proposed method CLIP-VIS in detail, which is built upon the encoder-decoder structure of Mask2Former \cite{cheng2022masked}. The encoder-decoder contains an image encoder, a pixel decoder, and a transformer decoder. To address the challenges of open-vocabulary video instance segmentation, we replace original image encoder with frozen CLIP \cite{radford2021learning} image encoder, fully levering its robust zero-shot classification ability. Instead of ViT-based CLIP, we employ a convolutional-based CLIP to extract multi-scale feature maps and retain spatial information by omitting the last pooling operation. Building upon CLIP based encoder-decoder, we introduce three modules, including class-agnostic mask generation, temporal top$K$-enhanced matching, and weighted open-vocabulary classification. These modules enable CLIP-VIS to effectively segment and track instances belonging to an open set of object categories within video sequences.

\noindent\textbf{Overall architecture.} Fig. \ref{overview} presents the overall architecture of our CLIP-VIS.  Given an input frame image within a video with the spatial size of $H\times W$, the feature map generated by CLIP image encoder is represented as $\mathcal{F}_{enc}^i\in \mathbb{R}^{{C}_{i} \times H/S_i\times W/S_i}$ and the output feature maps of pixel decoder are represented as $\mathcal{F}_{dec}^i \in \mathbb{R}^{256\times H/S_i\times W/S_i}, i=1,2,3,4$, where $S_1,S_2,S_3,S_4$ are respectively equal to 32, 16, 8, 4, and $C_i$ is the number of channels in the output of  \textit{i}-th level of  image encoder. In class-agnostic mask generation, we initialize a set of learnable queries, which are sent into the transformer decoder along with $\mathcal{F}_{dec}$ to obtain the query embeddings
$\mathcal{F}_{emb}\in \mathbb{R}^{N\times256}$, where $N$ is the number of  queries. Then, we multiply $\mathcal{F}_{dec}^4 \in \mathbb{R}^{256\times H/4\times W/4}$  with query embeddings $\mathcal{F}_{emb}\in \mathbb{R}^{N\times256}$ to obtain the mask $\mathcal{M}\in \mathbb{R}^{N\times H/4\times W/4}$.
We also employ query embeddings $\mathcal{F}_{emb}$ to predict  object scores $\mathcal{S}_{obj}\in \mathbb{R}^{N\times2}$ and mask IoU scores $\mathcal{S}_{iou}\in \mathbb{R}^{N\times1}$ through two different heads. The $\mathcal{S}_{obj}$ indicates whether the mask corresponds to an object and the $\mathcal{S}_{iou}$ represents the Intersection over Union (IoU) between the predicted mask and the ground-truth mask. Further, temporal top$K$-enhanced matching module selects $K$ mostly matched frames in memory and then performs query matching between current frame and these selected frames. In weighted open-vocabulary classification, we first employ mask pooling operation  to obtain visual embeddings $\mathcal{F}_{v} \in \mathbb{R}^{N\times D}$ from feature map $\mathcal{F}_{enc}^4$ according to the permuted masks $\mathcal{\hat M}$, where D is the dimensionality of visual and text embeddings. Then, we perform mask classification according to similarity scores between query visual features $\mathcal{F}_{v}$ and category text embeddings $\mathcal{F}_{t}\in \mathbb{R}^{L\times D}$ generated by CLIP text encoder, where $L$ represents the number of categories. Finally, we refine the classification scores  by incorporating the object scores $\mathcal{S}_{obj}$ and mask IoU scores $\mathcal{S}_{iou}$, further enhancing the correlation between mask prediction and classification.

\subsection {Class-Agnostic Mask Generation}

The class-agnostic mask generation module aims to predict query masks, and their corresponding object scores and mask IoU scores.  As shown in Fig. \ref{overview}, for a frame image in a video, the class-agnostic mask generation module utilizes CLIP image encoder and pixel decoder to extract features $\mathcal{F}_{enc}$ and $\mathcal{F}_{dec}$. Then, the transformer decoder predicts query embeddings $\mathcal{F}_{emb}$ using a set of learnable queries with the features $\mathcal{F}_{dec}$.
The predicted masks $\mathcal{M}$ are generated by  query embeddings $\mathcal{F}_{emb}$ and feature maps $\mathcal{F}_{dec}^4$ as
\begin{equation} 
\mathcal{M}  = \text{Sigmoid}(\mathcal{F}_{emb} \otimes \mathcal{F}_{dec}^4),
\end{equation}
where  $\otimes $ represents matrix multiplication.  To predict object scores and mask IoU scores of different queries, we attach two different branches, including object head and maskIoU head, to the query embeddings $\mathcal{F}_{emb}$. The object scores $\mathcal{S}_{obj}$ and mask IoU scores $\mathcal{S}_{iou}$ can be generated as
\begin{equation} 
\mathcal{S}_{obj}  = \text{Softmax}(\text{MLP}(\mathcal{F}_{emb})),
\end{equation}
\begin{equation} 
\mathcal{S}_{iou}  = \text{Sigmoid}(\text{MLP}(\mathcal{F}_{emb})).
\end{equation}
Each branch employs a multi-layer perceptron (MLP) consisting of three fully connected layers, with a Rectified Linear Unit (ReLU) activation function  after each layer.

Based on the predicted masks $\mathcal{M}$, we can generate the corresponding visual features of queries from CLIP image encoder for mask classification. In addition, we employ object scores and mask IoU scores to refine mask classification.

\begin{figure}[t]
    \centering
    \includegraphics[width=1.0\linewidth]{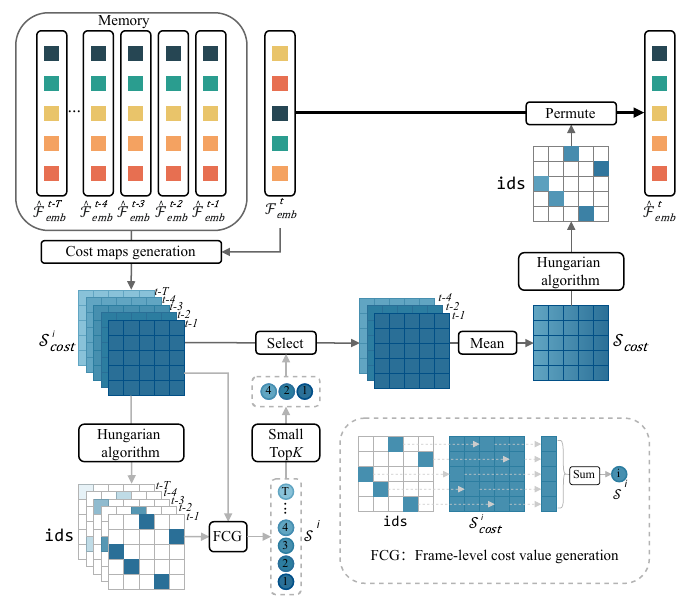}
    \caption{\textbf{Pipeline of temporal top$K$-enhanced matching.} We first select $K$ mostly matched frames from last $T$ frames according to frame-level cost values. Afterwards, we perform query matching using selected frames. In addition, we append the query embeddings in current frame in memory bank.}
    \label{match}
\end{figure}

\subsection {Temporal Top$K$-enhanced Matching}
To effectively track instances within a video, it is essential to associate query embeddings belonging to the same instance across frames. One common strategy as in MinVIS\cite{huang2022minvis} is to employ bipartite matching of query embeddings between two consecutive frames using Hungarian algorithm \cite{kuhn1955hungarian}. However this method easily propagates matching errors occurred at current frame to the following frames. In contrast, OV2Seg \cite{wang2023towards} aggregates query embeddings from all previous frames, and performs bipartite matching between the aggregated query embeddings and query embeddings at current frame. Although OV2Seg reduces matching error propagation, we argue that it remains susceptible to interference from noisy frames within the aggregated history. To overcome these limitations, we propose a temporal top$K$-enhanced matching strategy. Instead of using all previous frames, we select K mostly matched frames from previous short-term frames. Subsequently, we conduct bipartite matching based on the query embeddings from these K frames. This approach effectively reduces the impact of noisy frames and minimizes error propagation, leading to more robust and accurate instance tracking.

Fig. \ref{match} gives the pipeline of our temporal top$K$-enhanced matching strategy. We store the permuted query embeddings $\mathcal{\hat F}_{emb}^{t-T},\mathcal{\hat F}_{emb}^{t-T+1},...,\mathcal{\hat F}_{emb}^{t-1}$ from last $T$ frames in a memory bank, where  the features of same index in permuted query embeddings belong to same instance across frames. For current frame $t$,  we calculate cost maps $\mathcal{S}_{cost}^{i} \in \mathbb{R}^{ N\times N}, i =1,2,..,T,$ between query embeddings at current frame $t$ and permuted query embeddings from $T$ adjacent frames as 
\begin{equation} 
\mathcal{S}_{cost}^{i} = 1 - \cos(\mathcal{F}_{emb}^{t},\mathcal{\hat F}_{emb}^{t-i}), i =1,2,..,T.
\end{equation}
The large value in cost map $\mathcal{S}_{cost}^{i}$ represents small probability that the query embeddings from two frames belong to same instance. Based on these cost maps $\mathcal{S}_{cost}^{i}, i=1,2,...,T$, we employ the Hungarian algorithm to associate query embeddings from the current frame with permuted query embeddings from last $T$ frames stored in memory as
\begin{equation} 
\texttt{ids}^i = \text{Hungarian}(\mathcal{S}_{cost}^{i}),  i =1,2,..,T,
\end{equation}
where $\texttt{ids}^i \in \mathbb{R}^{N}$ can map the query embeddings at current frame $t$ to permuted  query embeddings at frame $t-i$. Then, we can generate frame-level cost value by aggregating cost values of matched query embeddings as
\begin{equation} 
s^i = \sum\limits_{n = 1}^N {\mathcal{S}_{cost}^{i}(n,\texttt{ids}^i(n))}, i =1,2,..,T.
\end{equation}
When $s^i$ is small, it represents that frame $t$ and frame $t-i$ has a high matching degree. Therefore, we select $K$ frames with minimum cost values, and mean their cost maps, which can be written as  
\begin{equation} 
\mathcal{S}_{cost} = \frac{1}{K}\sum (t-k) \mathcal{S}_{cost}^{k}, k \subseteq i^{k},
\end{equation}
where $i^{k}$ represent the index of selected $K$ frames.
Based on final cost map $\mathcal{S}_{cost}$, we can use Hungarian algorithm to associate queries at current frame with the matched queries in memory as 
\begin{equation} 
\texttt{ids} = \text{Hungarian}(\mathcal{S}_{cost}),
\end{equation}
where \texttt{ids} map the query embeddings at current frame to permuted  query embeddings in memory bank. In addition, we permute the query embeddings $\mathcal{F}_{emb}^t$ at current frame to generate permuted query embeddings $\mathcal{\hat F}_{emb}^{t}$ = $\mathcal{F}_{emb}^t(\texttt{ids})$ and store the permuted query embeddings $\mathcal{\hat F}_{emb}^{t}$ into  memory bank.

Our temporal top$K$-enhanced matching strategy can adaptively select mostly matched frames, and associate queries of current with previous $K$ mostly matched frames. Therefore, it can avoid error propagation and better deal with  complex scenarios, such as instance disappearing and reappearing.

\subsection {Weighted Open-Vocabulary Classification}
\label{sec:cls}
Our weighted open-vocabulary classification performs mask classification by fully utilizing robust zero-shot classification ability of frozen CLIP, where we calculate cosine similarity between visual and text embeddings of different queries. Instead of image-level classification, we perform instance-level classification by extracting query visual features from feature maps $\mathcal{F}_{enc}^4$. As shown in Fig. \ref{overview}, we generate query visual embeddings $\mathcal{F}_{v}\in \mathbb{R}^{N\times D}$ by mask pooling \cite{xu2023open,yu2024convolutions} as
\begin{equation} 
\mathcal{F}_{v}^{i}  = \frac{\text{sum}(\mathcal{F}_{enc}^4 \times (\mathcal{\hat M}^i > 0.5),\dim=(1,2))}{\text{sum}(\mathcal{\hat M}^i > 0.5),\dim=(1,2))}, i=1,2,...,N
\end{equation}
where $\dim=(1,2)$ represents the sum operation along spatial (height and width) dimension. Then, we employ CLIP text encoder to generate text embeddings of different categories, which are represented as $\mathcal{F}_{t}\in \mathbb{R}^{L\times D}$. Based on query visual embeddings $\mathcal{F}_{v}$ and text embeddings $\mathcal{F}_{t}$, we  calculate query classification scores $\mathcal{S}_{cls} \in \mathbb{R}^{N\times L}$ belonging to different categories using cosine similarity  as 
\begin{equation} 
\mathcal{S}_{cls}  = \cos(\mathcal{F}_{v}, \mathcal{F}_{t}).
\end{equation}

Although we assign each query with corresponding classification score, we can not ensure that the query with high classification score has high-quality mask prediction or belongs to the object, which will degrade performance. To reduce this negative effect, we introduce a weighting strategy to refine the classification scores as
\begin{equation} 
\mathcal{S}_{cls}^*  = \mathcal{S}_{cls}  \times S_{obj} \times S_{iou}.
\end{equation}
Then the categories of queries can be calculated as
\begin{equation} 
\mathcal{C}  = \arg\max(\mathcal{S}_{cls}^*, \dim=-1),
\end{equation}
where $\dim=-1$ represents $\arg\max$ operation along last dimension.

Compared to  original scores $\mathcal{S}_{cls}$, the refined scores $\mathcal{S}_{cls}^*$ calibrate classification scores with mask predictions, including the quality and object scores of predicted masks. As a result, the weighting strategy can decrease the misalignment between mask prediction and mask classification.

\subsection{Training Loss}

During training, we freeze CLIP image encoder and text encoder, and train pixel decoder, transformer decoder, and mask prediction head. In addition, our CLIP-VIS performs mask classification based on pre-trained CLIP, and performs query matching based on query embeddings. Therefore, our CLIP-VIS does not require category and identity annotations.
The overall training loss $\mathcal L$  can be written as 
\begin{equation}
\mathcal{L} = {\lambda_{obj}}{\mathcal{L}_{obj}}  
+ {\lambda_{iou}}{\mathcal{L}_{iou}}+ {\lambda_{mask}}{\mathcal{L}_{mask}}+ {\lambda _{dice}}{\mathcal{L}_{dice}},
\end{equation} 
where $\mathcal{L}_{obj}$ represents cross-entropy loss for the object head, and $\mathcal{L}_{iou}$ represents L1 loss in mask IoU branch. $\mathcal{L}_{mask}$ and $\mathcal{L}_{dice}$ represent binary cross-entropy and dice losses\cite{milletari2016v} for mask prediction following previous works \cite{huang2022minvis,wang2023towards,guo2023openvis}.  $\lambda_{obj},\lambda_{iou},\lambda_{mask},\lambda_{dice}$ are corresponding weighting factors, which are set as 2.0, 2.0, 5.0, and 5.0 respectively.

\section{Experiments}
In this section, we first introduce the datasets and evaluation metrics. Subsequently, we provide a detailed description of training and testing implementation details. Finally, we perform the experiments to demonstrate the effectiveness of our  CLIP-VIS and compare with state-of-the-art methods.

\begin{table*}[t!]
    \centering
	\renewcommand\tabcolsep{4.0pt}
	\renewcommand\arraystretch{1.1}
    \caption{\textbf{Comparison with state-of-the-art methods.}  `/' indicates that the training  and  validation sets belong to the same dataset with a fixed set of category, which is adopted by classical video instance segmentation methods. `*' represents the $AP$ for the uncommon class ($AP_u$) in BURST. \textsuperscript{$\dag$} indicates the backbone in additional CLIP.}
	\begin{tabular}{l|cc|ccccccccccc}
    \toprule
    \multirow{2}{*}{Method}  & \multirow{2}{*}{Training Data}& \multirow{2}{*}{Backbone} & \multicolumn{2}{c}{LV-VIS val.} & \multicolumn{2}{c}{LV-VIS test}& \multicolumn{1}{c}{OVIS} & \multicolumn{2}{c}{YTVIS2019} & \multicolumn{2}{c}{YTVIS2021}  & \multicolumn{2}{c}{BURST}  \\ 
    & \multicolumn{1}{c}{}& \multicolumn{1}{c|}{}& \multicolumn{1}{c}{$AP$} & \multicolumn{1}{c}{$AP_n$} & \multicolumn{1}{c}{$AP$} & \multicolumn{1}{c}{$AP_n$}& \multicolumn{1}{c}{$AP$} & \multicolumn{1}{c}{{$AP$}} & \multicolumn{1}{c}{$AP_n$}& \multicolumn{1}{c}{$AP$} & \multicolumn{1}{c}{$AP_n$}& \multicolumn{1}{c}{$AP$} & \multicolumn{1}{c}{$AP_n$}\\ 
    \midrule
    \multicolumn{12}{c}{\textcolor{gray!80}{\textit{Classical video instance segmentation}}} \\
    \midrule
    MaskTrack R-CNN\cite{yang2019video}      & / &R50 & - & - & - & - & 10.8 & 30.3 & - & 28.6 & - & - & -    \\
    SipMask\cite{cao2020sipmask}            & / &R50 & - & - & - & - & 10.2 & 33.7 & - & 31.7 & - & - & -   \\
    Mask2Former\cite{cheng2021mask2former}  & / &R50 & - & - & - & - & 17.3 & 46.4 & - & 40.6 & - & - & -    \\
    MinVIS\cite{huang2022minvis}            & / &R50 & - & - & - & - & 25.0 & 47.4 & - & 44.2 & - & - & -    \\
    IDOL\cite{wu2022defense}                & / &R50 & - & - & - & - & 30.2 & 49.5 & - & 43.9 & - & - & -    \\
    \midrule
    \multicolumn{12}{c}{\textcolor{gray!80}{\textit{Open-vocabulary video instance segmentation}}}   \\
    \midrule
    DetPro\cite{du2022learning}-SORT\cite{bewley2016simple}    & LVIS  &R50        & 6.4  & 3.5  & 5.8  & 2.1  & -    & -    & -    & -    & -   & -   &-     \\
    Detic\cite{zhou2022detecting}-SORT\cite{bewley2016simple}  & LVIS  &R50        & 6.5  & 3.4  & 5.7  & 2.1  & 6.7  & 14.6 & 3.5  & 12.7 & 3.1 & 1.9 & 2.5  \\
    DetPro\cite{du2022learning}-OWTB\cite{liu2022opening}      & LVIS  &R50        & 7.9  & 4.2  & 7.0  & 2.9  & -    & -    & -    & -    & -   & -   &-     \\
    Detic\cite{zhou2022detecting}-OWTB\cite{liu2022opening}    & LVIS  &R50        & 7.7  & 4.2  & 7.0  & 2.8  & 9.0  & 17.9 & 4.5  & 16.7 & 5.8 & 2.7 & 1.8  \\
    Detic\cite{zhou2022detecting}-XMem\cite{cheng2022xmem}     & LVIS  &R50        & 8.8  & 5.4  & 7.7  & 3.6 & -    & -    & -    & -    & -   & -   &-     \\
    OV2Seg\cite{wang2023towards}                               & LVIS  &R50        & 14.2 & 11.9 & 11.4 & 8.9  & 11.2 & 27.2 & 11.1 & 23.6 & 7.3 & 3.7 & 2.4  \\
    OVFormer\cite{fang2024unified}                                             &  LVIS  &  R50+ViT-B\textsuperscript{$\dag$}      &  - &  - & - & - & \textbf{15.1}&\textbf{34.8} & 16.5 & 29.8 & 15.7  & -  & - \\
    \textbf{CLIP-VIS(Ours)}                                    & LVIS  &R50        & \textbf{19.5} & \textbf{24.2} & \textbf{14.6} & \textbf{15.9} & 14.1 & 32.2 & \textbf{23.8} & \textbf{30.1} & \textbf{17.9} & \textbf{5.2} & \textbf{7.7}  \\
    \midrule
    Detic\cite{zhou2022detecting}-SORT\cite{bewley2016simple}  & LVIS  &SwinB      & 12.8 & 6.6  & 9.4  & 4.7  & 11.7 & 23.8 & 7.9  & 21.6 & 9.8  & 2.5  & 1.0 \\
    Detic\cite{zhou2022detecting}-OWTB\cite{liu2022opening}    & LVIS  &SwinB      & 14.5 & 8.5  & 11.8 & 6.1  & 13.6 & 30.0 & 9.7  & 27.1 & 11.4 & 3.9  & 2.4 \\
    Detic\cite{zhou2022detecting}-XMem\cite{cheng2022xmem}                                                 & LVIS  &SwinB      & 16.3 & 10.6 & 13.1 & 7.7  & -    & -    & -    & -    & -    & -    &-    \\
    OV2Seg\cite{wang2023towards}                               & LVIS  &SwinB      & 21.1 & 16.3 & 16.4 & 11.5 & 17.5 & 37.6 & 21.3 & 33.9 & 18.2 & 4.9  & 3.0 \\
    
    OVFormer\cite{fang2024unified}                                          & LVIS  &SwinB+ViT-B\textsuperscript{$\dag$}     & - & - & - & - & \textbf{21.3}&\textbf{44.3} &21.5 & 37.6 & 18.3  & -  & - \\
     \textbf{CLIP-VIS(Ours)}                                    & LVIS  &ConvNeXt-B        & \textbf{32.2} & \textbf{40.2} & \textbf{25.3} & \textbf{30.6} & 18.5 & 42.1 & \textbf{27.5} & \textbf{37.9} & \textbf{22.0} & \textbf{8.3} & \textbf{12.7}  \\
     \midrule
     \midrule
   {OpenVIS\cite{guo2023openvis}}  & {COCO,YT19} &{R50+R50\textsuperscript{$\dag$}}   & - & - & - & -& - &-             & - & - & - & {2.6}          & {$2.3^*$} \\
 {OpenVIS\cite{guo2023openvis}}  & {COCO,YT19} &{R50+ViT-B\textsuperscript{$\dag$}}  & - & -& - & - & - &{36.1}          & - & - & - & {3.5}         & {$3.0^*$} \\
    \textbf{CLIP-VIS(Ours)}       & COCO,YT19 &R50           & \textbf{9.4} &\textbf{13.7}  & \textbf{6.7} & \textbf{8.4} & \textbf{15.8} &\textbf{39.7} & - &\textbf{35.7}  &-  & \textbf{4.2} & \textbf{3.7$^*$}  \\
    \bottomrule
    \end{tabular}
    \label{results}
\end{table*}

\subsection {Datasets and Evaluation Metrics}

 We first train our CLIP-VIS on the training set of dataset LVIS\cite{gupta2019lvis}, or the training set of datasets COCO\cite{lin2014microsoft} and YTVIS2019\cite{yang2019video}. Afterwards, we evaluate the performance on both validation and test set of LV-VIS\cite{wang2023towards} dataset and validation set of multiple video instance segmentation datasets, including OVIS\cite{qi2022occluded}, YTVIS2019/2021\cite{yang2019video}, and BURST\cite{athar2023burst}. We give a brief introduction of these datasets as follows.

\noindent\textbf{LVIS} \cite{gupta2019lvis}  is a large-scale dataset designed for image open-vocabulary instance segmentation. There are over 1$k$ instance categories, 164$k$ images with about 2 million high-quality instance masks. The training set contains about 120$k$ images, which are used for network learning.

\noindent\textbf{COCO}\cite{lin2014microsoft} is a versatile dataset suitable for image detection, image segmentation, image captioning, and other tasks. It comprises 80 object categories, with a total of 330$k$ images and 1.5 million annotated objects. Similar to LVIS, it also has about 120$k$ images for training.

\noindent\textbf{LV-VIS}\cite{wang2023towards} is a large-scale video 
dataset for open-vocabulary video instance segmentation.  It contains 4828 videos and 1196 instance categories. The training set has 3083 videos, the validation set has 837 videos, and the test set has 980 videos.

\noindent\textbf{OVIS} \cite{qi2022occluded} is a dataset for occluded video instance segmentation. It contains 901 videos and 25 categories.  There are 5223 instances with well-annotated masks in each frame. The training set has 607 videos, the validation set has 140 videos, and the test set has 154 videos.

\noindent\textbf{YTVIS2019/2021} \cite{yang2019video} are two widely-used video instance segmentation dataset collected from video object segmentation dataset YouTube-VOS \cite{xu2018youtube}. YTVIS2019 is one of earliest video instance segmentation datasets, which contains 2883 videos and 40 categories. There are 4883 instances with annotated masks. The training set has 2238 videos, the validation set has 302 videos, and the test set has 343 videos. YTVIS2021 is an extension of YTVIS2019. It contains 3859 videos and 8171 instances belonging to 40 categories. The training set has 2985 videos, the validation set has 421 videos, and the test set has 453 videos. 

\noindent\textbf{BURST} \cite{athar2023burst} is a dataset designed for object segmentation in videos, which is built on tracking dataset TAO \cite{dave2020tao}. It contains 2914 videos and 16$k$ instances belonging to 482 categories, which are split into common and uncommon categories. The training set has 500 videos, the validation set has 993 videos, and the test set has 1421 videos.

\noindent\textbf{Evaluation Metrics.} Similar to existing video instance segmentation approaches \cite{yang2019video,huang2022minvis,wu2022defense}, we adopt the Average Precision $AP$ as evaluation metric. $AP$ is averaged over multiple IoU thresholds ranging from 0.5 to 0.95. To better evaluate open-vocabulary performance, we also report the $AP_n$, which only is averaged value of $AP$ on novel categories.

\subsection{Implementation Details}

We freeze  the vision-language model CLIP during training, and only train pixel decoder, transformer decoder, object head, and maskIoU Head. In CLIP, we adopt the pre-trained backbone R50\cite{he2016deep} and ConvNext-B\cite{liu2022convnet ,schuhmann2022laion} as image encoder. The structures of pixel decoder and transformer decoder are similar to Mask2Former \cite{cheng2022masked}. We train CLIP-VIS on four NVIDIA RTX 3090 GPUs, and adopt the  AdamW \cite{loshchilov2018fixing} for optimization. The input image is resized into 1024 $\times$ 1024 pixels during training.  When comparing with OV2Seg\cite{wang2023towards}, we adopt the similar settings to OV2Seg. Specifically, we train our CLIP-VIS on the set of common and frequent categories in LVIS\cite{gupta2019lvis} dataset with the batch size of 8 images. The initial learning rate is set as 1e-4. There are totally 30 epochs, and the learning rate is decreased at epoch 26 and 28 by a factor of 0.1. When comparing with OpenVIS\cite{guo2023openvis}, we adopt the similar settings to OpenVIS. Specifically, we initially train our method on the COCO \cite{lin2014microsoft} dataset and fine-tune it on YTVIS2019\cite{yang2019video}. The fine-tune learning rate is set as 5e-5. There are totally 6$k$ iterations, the learning rate is decreased at iteration 5.4$k$ by a factor of 0.1. For a fair comparison, we report the inference speed of our proposed method on a single NVIDIA A100 GPU similar to OV2Seg\cite{wang2023towards}.

\subsection{Comparison With Other Methods}
Here we compare our method with state-of-the-art methods on various video instance segmentation datasets in Table \ref{results}.

\noindent\textbf{On LV-VIS.} When using the backbone R50 on validation set of LV-VIS, our CLIP-VIS achieves an $AP$ score of 19.5\%, while OV2Seg has an $AP$ score of 14.2\%. Namely, our CLIP-VIS outperforms OV2Seg by 5.3\% in terms of $AP$. Moreover, our CLIP-VIS has large improvement on novel categories. Specifically, our CLIP-VIS achieves an $AP_n$ score of 24.2\%, while OV2Seg has an $AP_n$ score of 11.9\%. Therefore, our CLIP-VIS outperforms OV2Seg by 12.3\% on novel categories. When using the backbone ConvNext-B on validation set of LV-VIS, our CLIP-VIS outperforms OV2Seg by 11.1\% and 23.9\% in terms of $AP$ and $AP_n$. On test set of LV-VIS, our CLIP-VIS outperforms OV2Seg  outperforms OV2Seg by 8.9\% and 19.1\% in terms of $AP$ and $AP_n$ when using ConvNext-B. 

\noindent\textbf{On OVIS.} When using the backbone R50, our CLIP-VIS achieves an $AP$ score of 14.1\%, while OV2Seg obtains an $AP$ score of 11.2\%. Namely, our CLIP-VIS outperforms OV2Seg by 2.9\%. With the backbone ConvNext-B, our CLIP-VIS outperforms OV2Seg by 1.0\%.

\noindent\textbf{On YTVIS2019.} When using the backbone R50, our CLIP-VIS outperforms OV2Seg by 5.0\% and 12.7\% in terms of $AP$ and $AP_n$. With the backbone ConvNext-B,  our CLIP-VIS outperforms OV2Seg by 4.5\% and 6.2\% in terms of $AP$ and $AP_n$. Moreover, our method with ConvNext-B outperforms OVFormer with Swin-B+ViT-B  by 6.0\% in terms of $AP_n$. Furthermore, OpenVIS achieves an $AP$ score of 36.1\% with the backbone R50 and the ViT-B CLIP image encoder, while our CLIP-VIS outperforms OpenVIS by 3.6\% in terms of $AP$ with backbone R50.

\noindent\textbf{On YTVIS2021.} Similar to YTVIS2019, our CLIP-VIS has significant improvements on YTVIS2021. For example, our CLIP-VIS outperforms OV2Seg by 6.5\% and 10.6\% in terms of $AP$ and $AP_n$ when using R50. Our CLIP-VIS using ConvNext-B outperforms OVFormer using SwinB+ViT-B by 3.7\% in terms of $AP_n$. 

\noindent\textbf{On BURST.} All the methods have the low performance on challenging BURST. Nonetheless, our CLIP-VIS still outperforms other methods.When using the backbone R50, our CLIP-VIS outperforms OV2Seg by 1.5\% and 5.3\% in terms of $AP$ and $AP_n$ while our CLIP-VIS outperforms OpenVIS by 1.6\% and 1.4\% in terms of $AP$ and $AP_{u}$. $AP_{u}$ refers to the $AP$ for the uncommon class. When using ConvNext-B, our CLIP-VIS outperforms OV2Seg by 3.4\% and 9.7\% in terms of $AP$ and $AP_n$. 

\begin{table}
    \centering
	\renewcommand\tabcolsep{4.0pt}
	\renewcommand\arraystretch{1.1}
    \caption{\textbf{Impact of integrating different modules into baseline.} `WOC' indicates weighted open-vocabulary classification, and `TKM' indicates temporal top$K$-enhanced matching. The baseline adopts the adjacent matching in MinVIS. Our method has significant performance improvement.}
	\begin{tabular}{cc|c|ccccc}
    \toprule
    \multirow{2}{*}{WOC} & \multirow{2}{*}{TKM} &\multirow{2}{*}{Backbone} &\multicolumn{2}{c}{LV-VIS}  & \multicolumn{1}{c}{OVIS} & \multicolumn{2}{c}{BURST}  \\
    & \multicolumn{1}{c|}{} & \multicolumn{1}{c|}{} & \multicolumn{1}{c}{$AP$} & \multicolumn{1}{c}{$AP_n$} & \multicolumn{1}{c}{$AP$} & \multicolumn{1}{c}{$AP$} & \multicolumn{1}{c}{$AP_n$}\\ 
    \midrule

    \texttimes  & \texttimes    &\multirow{3}{*}{R50}   & 9.2            & 11.0            & 6.1            & 3.1           & 2.8   \\
    \checkmark  & \texttimes    &                       & 18.2           & 22.2          & 10.7           & 4.1           & 5.4 \\
    \checkmark  & \checkmark    &                       & \textbf{19.5}  & \textbf{24.2}  & \textbf{14.1}  & \textbf{5.2}  & \textbf{7.7}\\
    \bottomrule
    \end{tabular}
    \label{score and topk}
\end{table}

\begin{table}[t]
    \centering
	\renewcommand\tabcolsep{4.0pt}
	\renewcommand\arraystretch{1.1}
    \caption{\textbf{Impact of object scores $S_{obj}$ and mask IoU scores $S_{iou}$.} Both two scores can improve the performance.}

	\begin{tabular}{cc|c|ccccc}
    \toprule
    \multirow{2}{*}{$S_{obj}$}  &\multirow{2}{*}{$S_{iou}$}  &\multirow{2}{*}{Backbone}  &\multicolumn{2}{c}{LV-VIS}  &\multicolumn{1}{c}{OVIS}  &\multicolumn{2}{c}{BURST}  \\
    
    & \multicolumn{1}{c|}{} & \multicolumn{1}{c|}{} & \multicolumn{1}{c}{$AP$} & \multicolumn{1}{c}{$AP_n$} & \multicolumn{1}{c}{$AP$} & \multicolumn{1}{c}{$AP$} & \multicolumn{1}{c}{$AP_n$}\\ 
    \midrule

    \texttimes  & \texttimes    &\multirow{4}{*}{R50}  & 8.9           & 10.6         & 7.1            & 4.7          & 5.2      \\
    \texttimes  & \checkmark    &                      & 10.3           & 12.5           & 8.2            &5.1          & 6.0    \\
    \checkmark  & \texttimes    &                      & 18.7           & 23.0           & 13.8           & 5.1  & 7.6   \\
    \checkmark  & \checkmark    &                      & \textbf{19.5}  & \textbf{24.2}  & \textbf{14.1}  & \textbf{5.2}  & \textbf{7.7} \\
    \bottomrule
    \end{tabular}
    \label{Object score and maskiou}
\end{table}

\begin{table}[t]
\centering
\renewcommand\tabcolsep{3.5pt}
\renewcommand\arraystretch{1.1}

\caption{\textbf{Comparison of different designs in  mask IoU head.} In (a), we compare different features for mask IoU head, where $\mathcal{F}_{emb}$ represents the output embeddings of transformer decoder, and $\mathcal{F}_{emb}^{p}$ denotes the embeddings generated by pooling the output feature maps $\mathcal{F}_{dec}^4$ of pixel decoder using the predicted mask. In (b), we show the impact of different activation function for mask IoU head.}
\begin{tabular}{c|c|c|cccccc}
    \toprule
  \multirow{4}{*}{(a)} & \multirow{2}{*}{Input features} &\multirow{2}{*}{Backbone}&\multicolumn{2}{c}{LV-VIS}  &\multicolumn{1}{c}{OVIS}  &\multicolumn{2}{c}{BURST}  \\
  & \multicolumn{1}{c|}{}& \multicolumn{1}{c|}{} & \multicolumn{1}{c}{$AP$} & \multicolumn{1}{c}{$AP_n$} & \multicolumn{1}{c}{$AP$} & \multicolumn{1}{c}{$AP$} & \multicolumn{1}{c}{$AP_n$}\\
    \cmidrule{2-8}
   
   &  $\mathcal{F}_{emb}^{p}$ & \multirow{2}{*}{R50}& 19.2& 19.3& 13.6& \textbf{5.3}& 7.2\\
   & $\mathcal{F}_{emb}$ & & \textbf{19.5}& \textbf{24.2}& \textbf{14.1}& 5.2& \textbf{7.7}\\
    \midrule
 \multirow{4}{*}{(b)} & \multirow{2}{*} {Activation Function} &\multirow{2}{*}{Backbone}&\multicolumn{2}{c}{LV-VIS}  &\multicolumn{1}{c}{OVIS}  &\multicolumn{2}{c}{BURST}  \\
  & \multicolumn{1}{c|}{}& \multicolumn{1}{c|}{} & \multicolumn{1}{c}{$AP$} & \multicolumn{1}{c}{$AP_n$} & \multicolumn{1}{c}{$AP$} & \multicolumn{1}{c}{$AP$} & \multicolumn{1}{c}{$AP_n$}\\
    \cmidrule{2-8}

  &  ReLU & \multirow{2}{*}{R50}& 18.8& 22.8& 12.3& 5.1& 7.4\\
  &  Sigmoid && \textbf{19.5}&\textbf{ 24.2}& \textbf{14.1}& \textbf{5.2}& \textbf{7.7}\\
    \bottomrule
\end{tabular}
    
\label{Maskiouhead}
\end{table}

\begin{table}[t]
    \centering
	\renewcommand\tabcolsep{4.0pt}
	\renewcommand\arraystretch{1.1}
    \caption{\textbf{Ablation study on temporal top$K$-enhanced matching method.} `Adjacent' refers to the matching using adjacent frame in MinVIS, `Long-term' refers to the matching using all previous frames in OV2Seg, and `Top$K$' refers to our temporal top$K$-enhanced matching. In our method, `$T$' represents the size of the memory bank, while `$K$' represents the selected number of frames.}
	\begin{tabular}{l|c|cc|ccccc}
    \toprule
    \multirow{2}{*}{Strategy}  & \multirow{2}{*}{Backbone} & \multirow{2}{*}{$T$} & \multirow{2}{*}{$K$}& \multicolumn{2}{c}{LV-VIS}& \multicolumn{1}{c}{OVIS}& \multicolumn{2}{c}{BURST}   \\ 
     & & & \multicolumn{1}{c|}{}& \multicolumn{1}{c}{$AP$} & \multicolumn{1}{c}{$AP_n$} & \multicolumn{1}{c}{$AP$} & \multicolumn{1}{c}{$AP$} & \multicolumn{1}{c}{$AP_n$}\\ 
    \midrule
    Adjacent                  &\multirow{2}{*}{R50}   & -           & -           & 18.2           & 22.2           & 10.7            & 4.1           & 5.4     \\
    Long-term                 &                       & -           & -           & 18.7           & 23.0           & 12.1           & 4.7           & 6.3 \\
    \midrule
    \multirow{13}{*}{Top$K$}  &\multirow{13}{*}{R50}  & 3           & 1           & 18.4           & 23.0           & 10.9            & 4.5           & 6.0 \\
                              &                       & 5           & 1           & 18.6           & 23.3           & 10.9            & 4.6           & 6.1 \\
                              &                       & 5           & 3           & 19.3           & 24.2           & 13.7           & 4.9           & 7.3 \\
                              &                       & 7           & 1           & 18.6           & 23.3           & 11.1           & 4.6           & 5.7\\
                              &                       & 7           & 3           & 19.5           & 24.2           & 13.7          & 5.1           & 7.1\\
                              &                       & 7           & 5           & 19.1           & 23.7           & 12.4          & 5.4           & 7.4 \\
                              &                       & 9           & 1           & 18.7           & 23.4           & 10.7            & 4.7           & 5.7 \\
                              &                       & \textbf{9}           & \textbf{3}           & \textbf{19.5}           & \textbf{24.2}           & \textbf{14.1}           & \textbf{5.2}           & \textbf{7.7} \\
                              &                       & 9           & 5           & 19.3           & 23.9           & 13.2           & 5.4           & 7.6 \\
                              &                       & 9           & 7           & 19.2           & 23.7           & 12.4           & 5.4           & 7.3 \\
                              &                       & 11          & 3           & 19.5           & 24.2           & 13.1           & 5.2           & 6.9\\
                              &                       & 11          & 5           & 19.4           & 24.0           & 13.0          & 5.5           & 7.7\\
                              &                       & 11          & 7           & 19.3           & 23.8           & 12.0           & 5.4           & 6.5\\
    \bottomrule
    \end{tabular}
    \label{Post-processing matching module}
\end{table}

\begin{figure*}[t]
	\centering
	\includegraphics[width=0.97\textwidth]{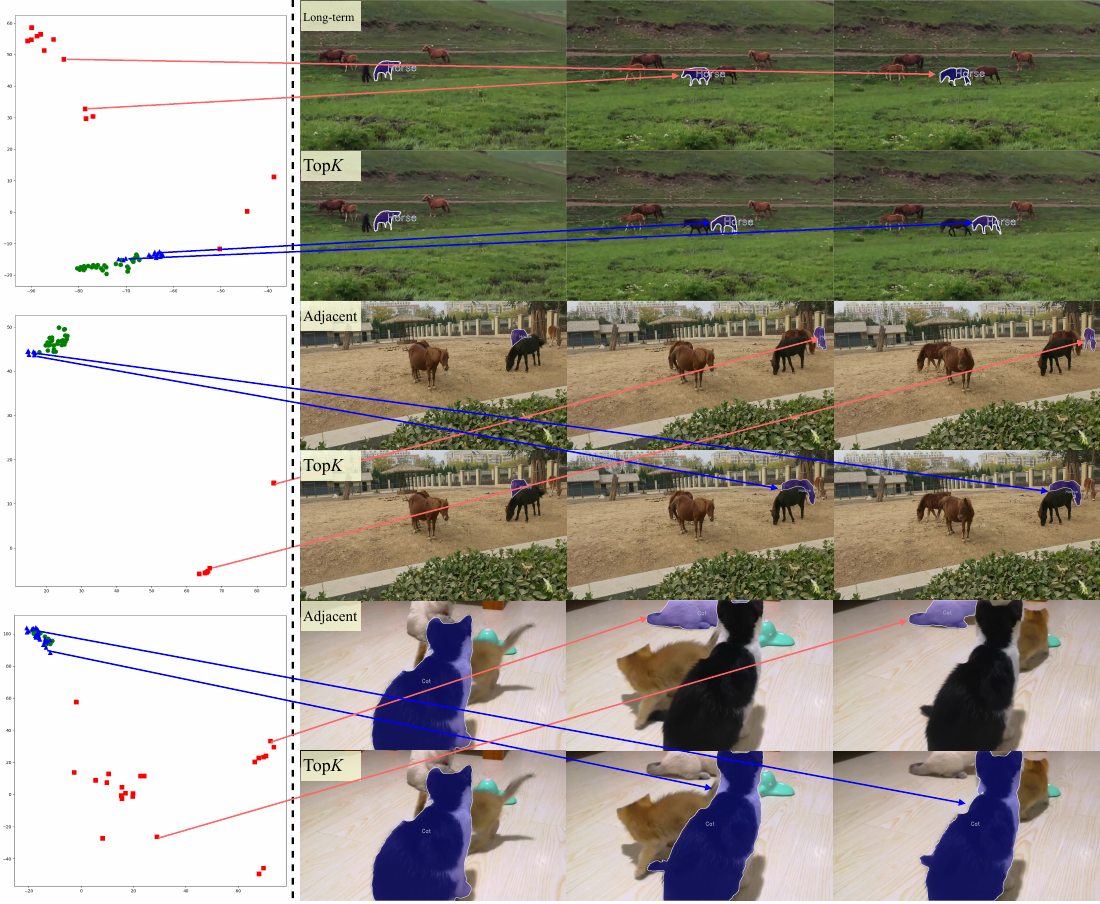} %
	\caption{\textbf{Visualizing query embeddings and qualitative results.} In left column, the points represent the query embeddings of an instance in all frames using t-SNE. Specifically, the circle points with green color represent the query embeddings with correct matching in both adjacent or long-term and top$K$ strategy, the triangle points with blue color represent the query embeddings with correct matching in our top$K$ strategy, and the square points with red color represent the query embeddings with wrong matching in adjacent or long-term strategy. In right column, we show  qualitative results corresponding to some points. The first row shows the results of adjacent or long-term strategy, and the second row shows the results of our top$K$ strategy.}
	\label{embed}
\end{figure*}

\subsection{Ablation Study}
Here we perform ablation study to show method effectiveness. We train our method on LVIS training set and evaluate it on validation set of LV-VIS, OVIS and BURST datasets.

\noindent\textbf{Impact of integrating different modules.} Table \ref{score and topk} shows the impact of gradually integrating weighted open-vocabulary classification and temporal top$K$-enhanced matching into the baseline. The baseline performs open-vocabulary classification without the weighting strategy, and associates the instances using previous adjacent frame as in MinVIS \cite{huang2022minvis}. On LV-VIS validation set, the baseline has the $AP$ score of 9.2\% and the $AP_n$ score of 11.0\%. When integrating weighted open-vocabulary classification (WOC) into the baseline, it has 9.0\% improvement on $AP$ and 11.2\% improvement on $AP_n$. When further integrating temporal top$K$-enhanced matching (TKM), it achieves the $AP$ score of 19.5\% and the $AP_n$ score of 24.2\%, outperforming the baseline by 10.3\% and 13.2\%. Similarly, in terms of $AP$, our CLIP-VIS outperforms the baseline by 8.0\% and 2.1\% on OVIS and BURST validation set.

\noindent\textbf{Impact of object scores and mask IoU scores.} Table \ref{Object score and maskiou} shows the impact of object scores $S_{obj}$ and mask IoU scores $S_{iou}$. When not using both object scores $S_{obj}$ and mask IoU scores $S_{iou}$, it has an $AP$ score of 8.9\% and 7.1\% on LV-VIS and OVIS, respectively. When only using the mask IoU scores $S_{iou}$, it has  the improvements of 1.4\% and 1.1\% in terms of $AP$ on LV-VIS and OVIS. With only object scores $S_{obj}$, it presents the improvements of 9.8\% and 6.7\% in terms of $AP$ on LV-VIS and OVIS. When using both scores, it obtains the $AP$ score of 19.5\% and 14.1\% on LV-VIS and OVIS, which has 10.6\% and 7.0\% improvements.

\noindent\textbf{Comparison of different design in mask IoU head.} First, we show the impact of the input features to mask IoU head. There are two different strategies to generate the input features for mask IoU head. The first one is directly using the query embeddings $\mathcal{F}_{emb}$ as input, while the second one is employing the predicted masks to extract the input features $\mathcal{F}_{emb}^p$ from the output feature map $\mathcal{F}_{dec}^4$ of pixel decoder with mask pooling. Table \ref{Maskiouhead}(a) present the results of these two strategies. We can observe that with the input features $\mathcal{F}_{emb}$, CLIP-VIS achieves higher $AP$ scores and $AP_n$ scores on different datasets, especially achieves higher $AP_n$ scores. For example, CLIP-VIS achieves an $AP_n$ score of 24.2\% with the query embeddings $\mathcal{F}_{emb}$ while an $AP_n$ score of 19.3\% with the pooling query embeddings $\mathcal{F}_{emb}^p$ on LV-VIS validation set. For simplicity and enhanced zero-shot classification ability, we employ the query embeddings $\mathcal{F}_{emb}$ as the input to mask IoU head. Then, we show the impact of different activation functions to mask IoU head, including ReLU and Sigmoid, in Table \ref{Maskiouhead}(b).  It has higher $AP$ scores and $AP_n$ scores using Sigmoid in mask IoU head.

\begin{figure*}[t!]
\centering
\captionsetup[subfloat]{labelfont=footnotesize,textfont=footnotesize}
\subfloat[{Results on novel category}]{\includegraphics[width=0.48\textwidth]{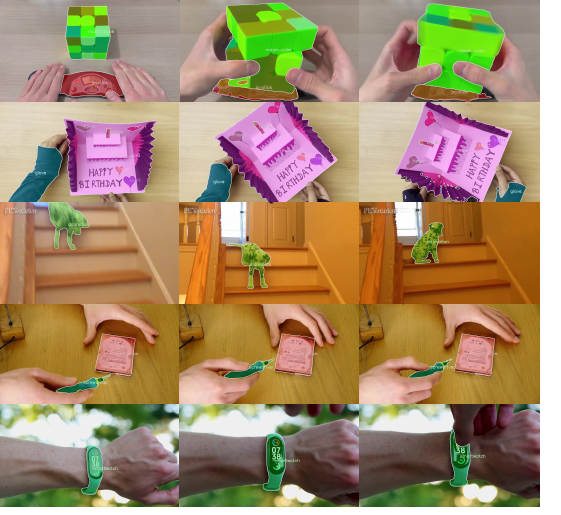}%
\label{Novel category results}}
\hfil
\subfloat[Results on similar category]{\includegraphics[width=0.48\textwidth]{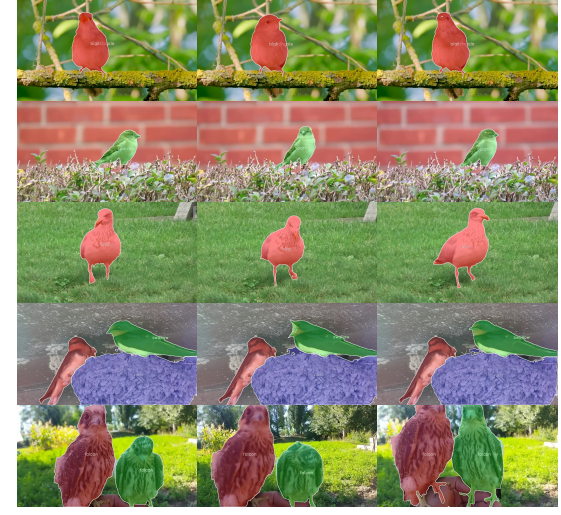}%
\label{Similar category results}}
\vfil
\vspace{-1.0em}
\subfloat[Results on complex scenes]{\includegraphics[width=0.97\textwidth]{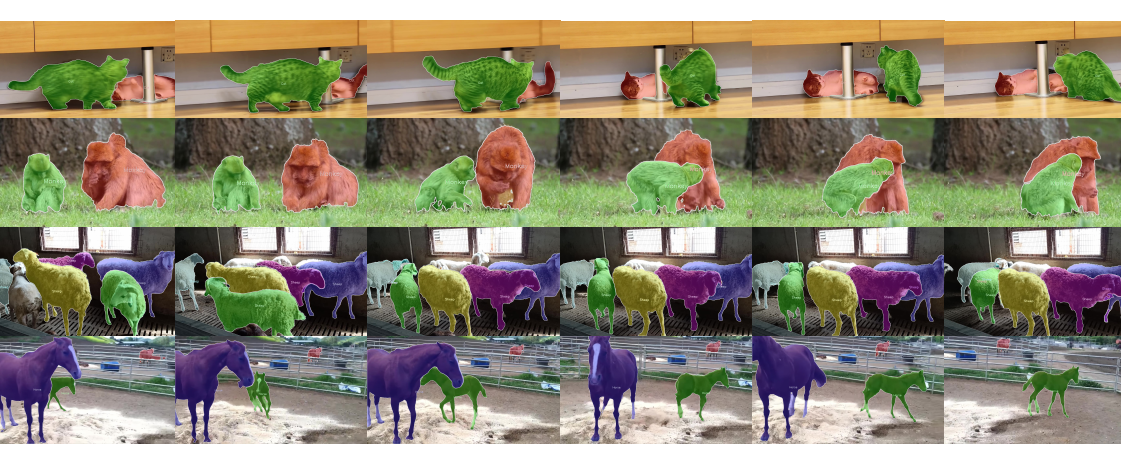}%
\label{Tracking results}}
\caption{\textbf{Qualitative results of our proposed method.} \textit{a)} shows some video instance segmentation results on novel categories. \textit{b)} displays some video instance  segmentation results of various bird species, where our method can accurately classify similar categories. \textit{c)} shows video instance  results in complex scenes. }
\label{visual}
\end{figure*}

\noindent\textbf{Impact of temporal top$K$-enhanced matching.} The top row of Table \ref{Post-processing matching module}  compares our temporal top$K$-enhanced matching with the related matching strategies used in MinVIS and OV2Seg. MinVIS performs query matching between current frame and previous one frame. We call it as adjacent matching.  OV2Seg performs query matching between current frame and aggregated frame from all previous the frames. We call it as long-term matching. The adjacent matching and long-term matching have the $AP$ scores of 18.2\% and 18.7\% on LV-VIS, while our temporal top$K$-enhanced matching strategy has the $AP$ score of 19.5\% on LV-VIS. Therefore, our top$K$ matching strategy outperforms adjacent and long-term matching by 1.3\% and 0.8\%. Furthermore,on the OVIS dataset, where videos are longer and scenes are more complex, our matching method achieves a greater performance improvement, outperforming adjacent and long-term matching by 3.4\% and 2.0\%, respectively. The bottom row of Table \ref{Post-processing matching module} further shows the impact of hyper-parameters $T,K$ in our top$K$-enhanced matching strategy. We observe that $T=9,K=3$ has almost best performance on LV-VIS, OVIS and BURST. 

We further visualize the query embeddings and some qualitative results of adjacent matching, long-term matching and our top$K$ matching in Fig. \ref{embed}. We only select one instance and show query embeddings  using the t-SNE\cite{van2008visualizing}. The first example presents the comparison results between long-term matching and our top$K$ matching strategy while the second and third examples depict the comparison results between adjacent matching and top$K$ matching strategy. In first column of each example, the points represent the query embeddings of the instance in all frames. All matching strategy share the same query embeddings, but generate different matching results represented as different colors. The circle points with green color represent the query embeddings with correct matching in  adjacent, long-term, and our top$K$ strategy, the triangle points with blue color represent the query embeddings with correct matching in our top$K$ strategy, and the square points with red color represent the query embeddings with wrong matching in adjacent or long-term strategy. In right column of each example, we show  the qualitative results corresponding to the different points, where the top show the results of adjacent or long-term strategy, and the bottom shows the results of our top$K$ strategy. It can observe that adjacent and long-term strategies can lead to tracking errors among similar instances while our top$K$ strategy can track the instance accurately and remove the wrong matching (square point).

\subsection{Qualitative Results}

\begin{figure}[tb]
\centering
\includegraphics[width=\linewidth]{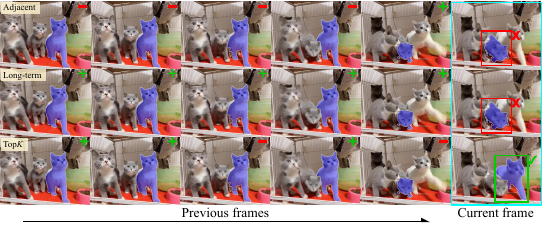} %
\caption{\textbf{Visualization of  frames used for matching.} The selected frames for matching are marked with `+', while the unused frames are marked with `-'.}
\label{vis1}
\end{figure}

Fig \ref{visual} presents some qualitative results of our proposed method. In Fig. \ref{Novel category results}, we show some video instance segmentation results on novel categories. Our CLIP-VIS can accurately classify these novel categories, such as hard drive, screwdriver, birthday card, dalmatian, magic cube and smartwatch. Moreover, our CLIP-VIS is capable of distinguishing between closely related categories, such as accurately classifying different species of bird, such as nightingale, sparrow, seagull, swallow and falcon in Fig. \ref{Similar category results}. In Fig. \ref{Tracking results}, we show the video instance segmentation results in complex scenes, including instance disappearing and reappearing, and instance occlusion. Our CLIP-VIS can track the instances in these complex scenes. Fig \ref{vis1} further visualizes the example of  frames used for matching.  Compared to adjacent or long-term matching, our method avoids selecting previous frame that exists a mistaken matching. As a result, our method perform a correct matching for the right cat in current frame.

\begin{table}[tb]
\centering
\renewcommand\tabcolsep{4.0pt}
\renewcommand\arraystretch{1.1}
\caption{\textbf{Comparison in terms of accuracy, FLOPs, and inference speed on LV-VIS.} Inference speed (FPS) of all methods are reported on a single NVIDIA A100. $^*$ indicates that OV2Seg is re-implemented by us for  fair comparison. Our method has the best trade-off in terms of accuracy and computational cost.}
\begin{tabular}{l|c|ccc}
\toprule
{Method}  & {Backbone} & {AP} & {FLOPs}& {FPS} \\ 

\midrule 
Detic\cite{zhou2022detecting}-SORT\cite{bewley2016simple}    &  R50        & 6.5  & -& 6.0  \\
Detic\cite{zhou2022detecting}-OWTB\cite{liu2022opening}      &  R50      & 7.7 & - & 5.9  \\
 Detic\cite{zhou2022detecting}-XMem\cite{cheng2022xmem}       &  R50        & 8.8 & - & 16.4 \\
OV2Seg\cite{wang2023towards}                                 &  R50       & 14.2 & - & 20.1  \\
 OV2Seg$^*$\cite{wang2023towards}                             & R50   &14.2    & \textbf{238.2G}  & 26.1 \\
\textbf{CLIP-VIS(Ours)}                                      &R50     &\textbf{19.5}  & 244.1G & \textbf{31.4}   \\
\midrule

Detic\cite{zhou2022detecting}-SORT\cite{bewley2016simple}  &SwinB     &12.8 & -& 6.7  \\
Detic\cite{zhou2022detecting}-OWTB\cite{liu2022opening}     &SwinB      &14.5  & -& 6.8  \\
 Detic\cite{zhou2022detecting}-XMem\cite{cheng2022xmem} &SwinB      &16.3  & - & 13.4\\
OV2Seg\cite{wang2023towards}                                 &  SwinB      & 21.1  & - & 16.8  \\
OV2Seg$^*$\cite{wang2023towards}                                 & SwinB     & 21.1 & 448.2G & 17.2   \\
\textbf{CLIP-VIS(Ours)}                                      & ConvNeXt-B     & \textbf{32.2}  & \textbf{409.3G}  & \textbf{21.0} \\
\bottomrule
\end{tabular}
\label{Complexity}
\end{table}

\subsection{Complexity Comparison}
Table \ref{Complexity} further presents the comparison in terms of accuracy, FLOPs, and inference speed on LV-VIS validation set.  It can seen that, our proposed method has the best trade-off between accuracy and computational cost. For instance, when using the same backbone R50, compared to OV2Seg \cite{wang2023towards}, our method is 5.3\% better in terms of AP, while operating at 1.20$\times$ faster speed. Moreover, our method with ConvNext-B is 11.1\% better than OV2Seg with SwinB in terms of AP, while operating at 1.22$\times$ faster speed.

\section{Conclusion}
We introduced a novel open-vocabulary video instance segmentation method, called CLIP-VIS. Our CLIP-VIS extends the frozen CLIP by introducing three modules, including class-agnostic mask generation, temporal top$K$-enhanced matching, and weighted open-vocabulary classification. Class-agnostic mask generation first predicts the query masks. Temporal top$K$-enhanced matching performs accurate query matching across frames. Based on the predicted masks, weighted open-vocabulary classification perform classification by considering the correlation between mask prediction and classification. Currently, our proposed method utilizes a pre-trained CLIP model to perform classification, which is trained at the image-level classification dataset. There exists some inconsistency between image-level and instance-level classification. In the future, we will continue our research to address these issues.

\bibliographystyle{IEEEtran}
\bibliography{Bibliography}

\begin{IEEEbiography}[{\includegraphics[width=1in,height=1.25in,clip,keepaspectratio]{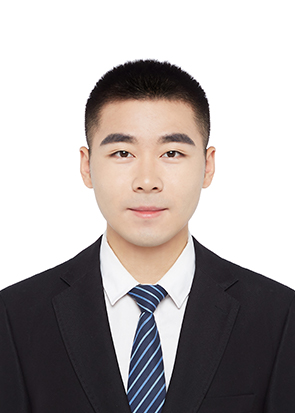}}]{Wenqi Zhu} received the B.E degree in electronic information engineering from Tianjin University, Tianjin, China, in 2023. He is currently a master student in Tianjin University. His research interests include object segmentation and image analysis.
\end{IEEEbiography}

\begin{IEEEbiography}[{\includegraphics[width=1in,height=1.25in,clip,keepaspectratio]{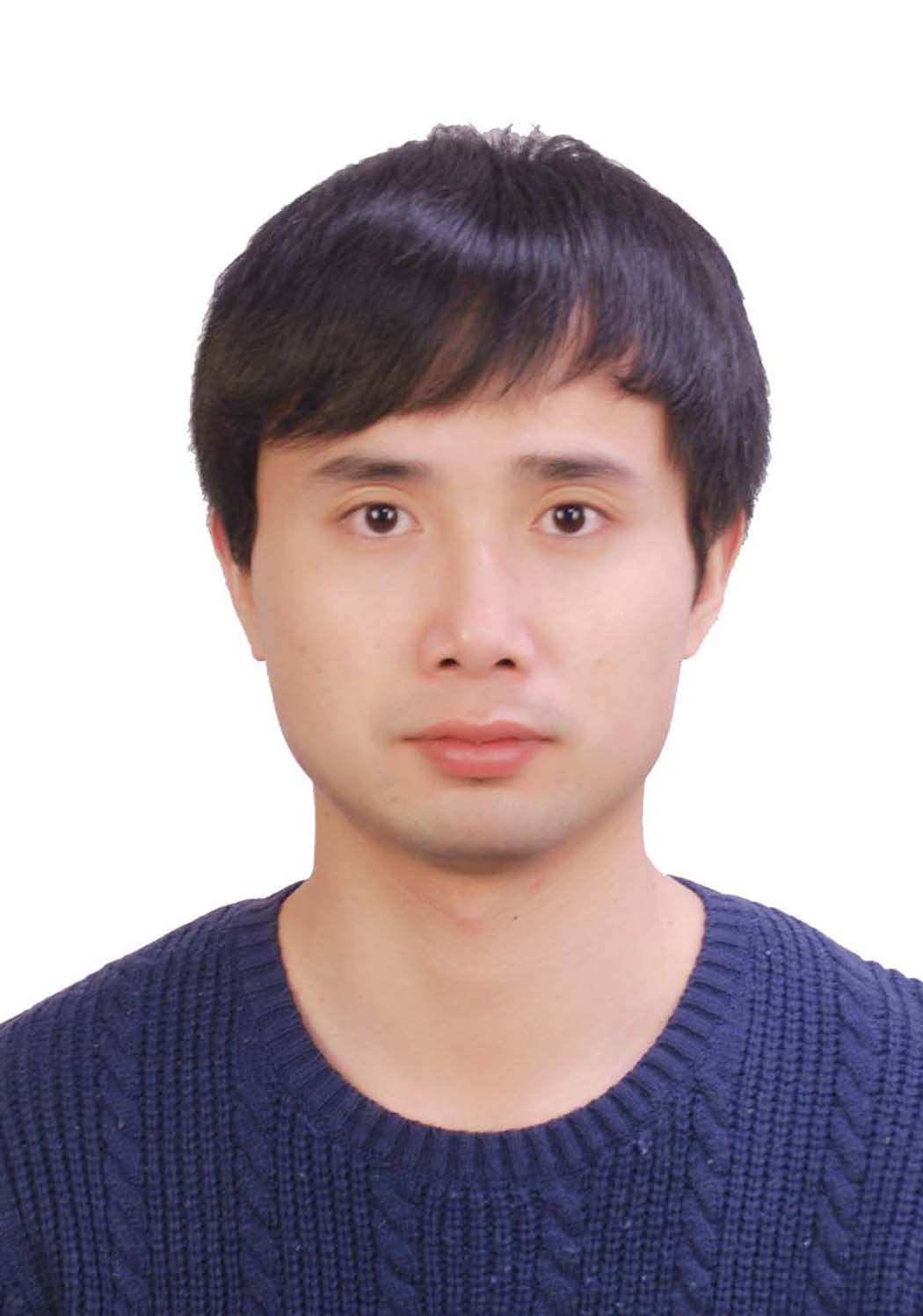}}]{Jiale Cao}(Member, IEEE) received the Ph.D degree in information and communication engineering from Tianjin University, Tianjin, China, in 2018. He is currently an Associate Professor with Tianjin University. His research interests include object detection and image analysis, in which he has published 30+ IEEE Transactions and CVPR/ICCV/ECCV articles. He serves as a regular Program Committee Member for leading computer vision and artificial intelligence conferences, such as CVPR, ICCV, and ECCV.
\end{IEEEbiography}
\
\begin{IEEEbiography}[{\includegraphics[width=1in,height=1.25in,clip,keepaspectratio]{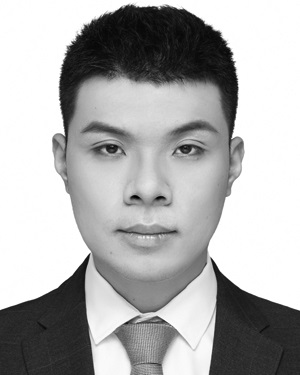}}]{Jin Xie} completed his Ph.D. in information and communication engineering with Tianjin University, Tianjin, China, in 2021. He currently serves as an Associate Professor at Chongqing University, Chongqing, China. His research interests include machine learning and computer vision, in which he has published more than 20 papers in various journals.
\end{IEEEbiography}

\begin{IEEEbiography}[{\includegraphics[width=1in,height=1.25in,clip,keepaspectratio]{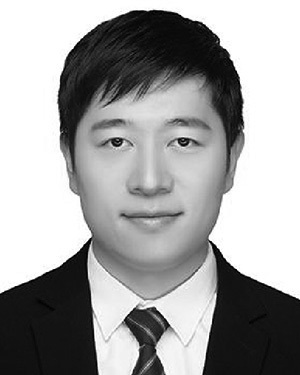}}]{Shuangming Yang}(Member, IEEE) received the Ph.D. degree in control science and engineering from Tianjin University, Tianjin, China, in 2020.,He was a Lecturer with Tianjin University from 2020 to 2021, where he is currently an Associate Professor with the School of Electrical and Information Engineering. He is also with Shanghai Artificial Intelligence Laboratory, Shanghai, China. His research interests include artificial general intelligence, neuromorphic computing, brain-inspired computing, and computer vision.
\end{IEEEbiography}

\begin{IEEEbiography}[{\includegraphics[width=1in,height=1.25in,clip,keepaspectratio]{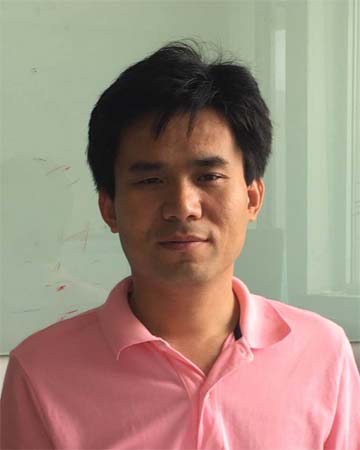}}]{Yanwei Pang}(Senior Member, IEEE) received the Ph.D. degree in electronic engineering from the University of Science and Technology of China in 2004. He is currently a Professor with Tianjin University, China, where he is also the Founding Director of the Tianjin Key Laboratory of Brain Inspired Intelligence Technology (BIIT). His research interests include object detection and image recognition, in which he has published 150 scientific papers, including 40 IEEE Transactions and 30 top conferences (e.g., CVPR, ICCV, and ECCV) papers.
\end{IEEEbiography}

\end{document}